\documentclass[preprint,12pt]{elsarticle}

\usepackage{lineno}
\usepackage{atbegshi}
\usepackage{graphicx}
\usepackage{hyperref}
\usepackage{epsfig}
\usepackage{rotating}
\usepackage{amssymb}
\usepackage{url}
\usepackage{array}
\usepackage{color}
\usepackage{amsmath}
\usepackage{makecell}
\usepackage{pdfpages}
\usepackage{tabularx}
\usepackage{multirow}

\usepackage[singlelinecheck=false,justification=justified 
]{caption}
\usepackage{diagbox}
\newcommand{\minisection}[1]{\vspace{0.04in} \noindent {\bf #1}\ \ }

\journal{CVIU}

\begin{document}

\begin{frontmatter}

 \title{Context Proposals for Saliency Detection\tnoteref{label1}}

 \author[rvt,focal]{Aymen~Azaza\corref{cor1}}
\ead{aymen.azaza@cvc.uab.es}  

\cortext[cor1]{Corresponding author}
 \address[rvt]{Noccs, National Engineering School of Sousse,
Tunisia.}

\author[focal]{Joost~van de Weijer\corref{cor1}}
\ead{joost@cvc.uab.es}
 \address[focal]{Computer Vision Center Barcelona, Spain.}

\author[rvt]{Ali~Douik}
\ead{ali.douik@enim.rnu.tn}

\author[focal]{Marc~Masana}
\ead{mmasana@cvc.uab.es}

\begin{abstract}
One of the fundamental properties of a salient object region is its contrast with the immediate context. The problem is that numerous object regions exist which potentially can all be salient. One way to prevent an exhaustive search over all object regions is by using object proposal algorithms. These return a limited set of regions which are most likely to contain an object. Several saliency estimation methods have used object proposals. However, they focus on the saliency of the proposal only, and the importance of its immediate context has not been evaluated.

In this paper, we aim to improve salient object detection. Therefore, we extend object proposal methods with context proposals, which allow to incorporate the immediate context in the saliency computation. We propose several saliency features which are computed from the context proposals. In the experiments, we evaluate five object proposal methods for the task of saliency segmentation, and find that Multiscale Combinatorial Grouping outperforms the others. Furthermore, experiments show that the proposed context features improve performance, and that our method matches results on the FT datasets and obtains competitive results on three other datasets (PASCAL-S, MSRA-B and ECSSD).
\end{abstract}


\end{frontmatter}

\section{Introduction}\label{sec:introduction}
To rapidly extract important information from a scene, the human visual system allocates more attention to salient regions. Research on computational saliency focuses on designing algorithms which, similarly to human vision, predict which regions in a scene are salient. In computer vision, saliency is used both to refer to eye-fixation prediction~\citep{wang2015atypical,xu2015turkergaze} as well as to salient object segmentation~\citep{Jiang2013salient,li2014secrets}. It is the latter which is the focus of this article. Computational saliency has been used in applications such as image thumbnailing~\citep{marchesotti2009framework}, compression~\citep{stella2009image}, and image retrieval~\citep{wan2009approach}.

Object proposal methods have recently been introduced in saliency detection methods~\citep{li2014secrets}. They were first proposed for object recognition, which was long dominated by sliding window approaches (see e.g.~\cite{felzenszwalb2010object}). Object proposal methods reduce the number of candidate regions when compared to sliding window approaches~\citep{uijlings2013selective}. They propose either a set of bounding boxes or image segments, which have a high probability of containing an object~\citep{hosang2015makes,Krahenbuhl2014Geodesic}. Recently, these methods have been applied in saliency detection~\citep{frintrop2015traditional,li2014secrets,wang2015deep}. Object proposals especially help in obtaining exact boundaries of the salient objects~\citep{li2014secrets}. In addition, they can reduce the computational costs of evaluating saliency based on a sliding window~\citep{liu2011learning}.

The saliency of an object is dependent on its context, i.e.~an object is salient (or not) with respect to its context. If a visual feature, e.g. color, textures or orientation, of an object differs from that of its context it is considered salient. Traditionally, this has been modeled in saliency computation with the center-surround mechanism~\citep{gaborski2003goal, han2012biological}, which approximates visual neurons. This mechanism divides the receptive field of neurons into two regions, namely the center and surround, thereby modeling the two primary types of ganglion cells in the retina. The first type is excited by a region in the center, and inhibited by a surround. The second type has the opposite arrangement and is excited from the surround and inhibited by a center. 
In computational saliency the center-surround mechanism has been implemented in different ways. For example, ~\cite{itti1998model} model this by taking the difference between fine (center) and coarse scale (surround) representations of image features. Even though this has been shown to successfully model eye fixation data, for the task of salient object detection this approach is limited to the shapes of the filters used. It can only consider the differences between circle regions of different radii. This led~\cite{liu2011learning} to consider center-surround between arbitrary rectangles in the images for salient object detection. In this work we will further generalize the concept of center-surround but now to arbitrarily shaped object proposals. 

To generalize the concept of center-surround to arbitrary shaped object proposals we extend object proposals with context proposals. We consider any object proposal method which computes segmentation masks. For each object proposal we compute a context proposal which encompasses the object proposal and indicates the part of the image which describes its direct surrounding. To compute the saliency with respect to the context proposals, we use a similar approach as~\cite{mairon2014closer}. For an object to be salient, it should be so with respect to the region described by the context proposal. In addition, because typically an object is occluding a background, it is expected that the features in the context proposal do not vary significantly. As a consequence, the saliency of the object proposal is increased if the corresponding context-proposal is homogeneous in itself, and different with respect to the object segment. In~\citep{mairon2014closer} these observations on context-based saliency led to an iterative multi-scale accumulation procedure to compute the saliency maps. Here, however, we circumvent this iterative process by directly computing context proposals derived from object proposals, and subsequently computing the context saliency between the proposal and its context proposal. 

The main contribution of our paper is that we propose several context based features for saliency estimation. These are computed from context proposals which are computed from object proposals. To validate our approach we perform experiments on a number of benchmark datasets. We show that our method matches state-of-the-art on the FT dataset and improves state-of-the-art results on three benchmark (PASCAL-S, MSRA-B and ECSSD datasets). In addition, we evaluate several off-the-shelf deep features and object proposal methods for saliency detection and find that VGG-19 features and multiscale combinatorial grouping (MCG) obtain the best performance.

This paper is organized as follows. In Section~\ref{sec:related work} we discuss the related work. In Section~\ref{sec:overview} we provide an overview of our approach to saliency detection. In Section~\ref{sec:context} the computation of context proposals is outlined. Next we provide details on the experimental setup in Section~\ref{sec:Exp_setup} and give results in Section~\ref{sec:Exp}. Conclusions are provided in Section~\ref{sec:concl}.


\section{Related work}\label{sec:related work}
In this section we provide an overview of salient object detection methods and their connection with object proposal methods. More complete reviews on saliency can be found in~\citep{borji2013state,zhao2013learning,Zhang2016areview}. 

\minisection{Saliency detection}
One of the first methods for computational saliency was proposed by ~\cite{itti1998model}. Their model based on the feature integration theory of ~\cite{treisman1980feature} and the work of ~\cite{Koch1987Shifts}  decomposes the input image into low level feature maps including color, intensity and orientation. These maps are subsequently merged together using linear filtering and center surround structures to form a final saliency map. Their seminal work initiated much research in biologically inspired saliency models~\citep{gaborski2003goal,murray2013low,siagian2007rapid} as well as more mathematical models for computational saliency~\citep{Achanta2009,harel2006graph,hou2007saliency,li2015adaptive}.  The central surround allows to measure contrast with the context, however it is confined to predefined shapes; normally the circle shape of the Gaussian filters~\citep{itti1998model} or rectangle shapes in the work of ~\cite{liu2011learning}. In this paper we will propose a method for arbitrary shaped contexts.

Local and global approaches for visual saliency can be classified in the category of bottom-up approaches. Local approaches compute local center-surround contrast and rarity of a region over its neighborhoods.~\cite{itti1998model} derive a bottom-up visual saliency based on center surround difference through multiscale image features. ~\cite{liu2011learning} propose a binary saliency estimation method by training a
CRF to combine a set of local, regional, and global  features.~\cite{harel2006graph} propose the GBVS method which is a bottom-up saliency approach that consists of two steps: the generation of feature channels as in Itti's approach, and their normalization using  a graph based approach. A saliency model that computes local descriptors from a given image in order to measure the similarity of a pixel to its neighborhoods was proposed by~\cite{seo2009static}.~\cite{garcia2009decorrelation} propose a AWS method which is based on the decorrelation and the distinctiveness of local responses. 

Another class of features for saliency are based on global context or rarity; the saliency of a feature is based on its rarity with respect to the whole image.~\cite{goferman2012context} consider the difference of patches with all other patches in the image to compute global saliency.~\cite{wang2011image} compute saliency by considering the reconstruction error which is left after reconstructing a patch from other patches (other patches can be from the same image or from the whole dataset).~\cite{Jiang2013salient} compute the rarity of a feature by comparing the contrast between a 15 pixel border around the image and the object proposal histogram.~Other than these methods we propose a method to compute the saliency with respect to the direct context of the object. Finally, to compute saliency~\cite{huo2016object} combined local and global objectness cues with a set of  candidates location.

Our work has been inspired by a recent paper~\citep{mairon2014closer} which demonstrates the importance of visual context for saliency computation. The work is based on the observation that an object is salient with respect to its context. And since context is an integral part of saliency of an object, it should therefore be assigned a prominent role in its computation. The final saliency map is computed by alternating between two steps: 1. the fusing of image regions based on their color distance into larger and larger context segments, and 2. the accumulation of saliency votes by the context segments (votes are casted to the region which is enclosed by the context segments). The steps are alternated until the whole image is clustered together into a single context segment. The procedure is elegant in its simplicity and was shown to obtain excellent results. However, the iterative nature of the computation renders it computationally very demanding.

Deep convolutional neural networks have revolutionized computer vision over the last few years.~This has recently led to several papers on deep learning for saliency detection~\citep{wang2015deep,li2015visual,zhao2015saliency,pan2016shallow,de2017saliency}. Both \cite{li2015visual} and ~\cite{zhao2015saliency} consider parallel networks which evaluate the image at various scales.~\cite{wang2015deep} use two networks to describe local and global saliency.~\cite{tong2015salient} combine a local and global model to compute saliency. The main challenge for saliency detection with deep networks is the amount of training data which is not always available. This is solved in~\citep{wang2015deep,li2015visual,zhao2015saliency} by training on the largest available saliency dataset, namely MSRA-B~\citep{liu2011learning}, and testing on the other datasets (both~\citep{li2015visual,Li2016deep} also use pretrained network weights trained on the 1M Imagenet dataset). Like these method, we will use a pretrained network for the extraction of features for saliency detection.

\minisection{Object Proposal methods}
Object detection based on object proposals methods has won in popularity in recent years~\citep{uijlings2013selective}. The main advantages of these methods is that they are not restricted to fixed aspect ratios as most sliding window methods are, and more importantly, they allow to evaluate a limited number of windows. As a consequence more complicated features and classifiers can be applied, resulting in state-of-the-art object detection results. The generation of object hypotheses can be divided into methods whose output is an image window and those that generate object or segment proposals.~The latter are of importance for salient object detection since we aim to segment the salient objects from the background. 

Among the first object proposal methods the work of~\cite{carreira2010constrained}, named the Constrained Parametric Min-Cuts (CPMC) method, uses graph cuts with different random seeds to obtain multiple binary foreground and background segments.~\cite{alexe2012measuring} proposes to measure the objectness of an image window, where they rank randomly sampled image windows based on their likelihood of containing the object by using multiple cues among which edges density, multiscale saliency, superpixels straddling and color contrast.~\cite{endres2014category} proposed an object proposal method similar to the CPMC method by generating multiple foreground and background segmentations. A very fast method for object proposals was proposed by \cite{cheng2014bing}, which generates box proposals at 300 images per second.

An extensive comparison of object proposal methods was performed by ~\cite{hosang2015makes}. Among the best evaluated object proposal methods (which generate object segmentation) are the selective search~\citep{uijlings2013selective}, the geodesic object proposals~\citep{Krahenbuhl2014Geodesic} and the multiscale combinatorial grouping  method~\citep{arbelaez2014multiscale}. Selective search proposes a set of segments based on hierarchical segmentations of the image where the underlying distance measures and color spaces are varied to yield a large variety of segmentations.~\cite{Krahenbuhl2014Geodesic}, propose the geodesic object proposals method, which applies a geodesic distance transfer to compute object proposals. Finally,  Multiscale Combinatorial Grouping~\citep{arbelaez2014multiscale} is based on a bottom-up hierarchical image segmentation. Object candidates are generated by a grouping procedure which is based on edge strength.

Several methods have applied object proposals to saliency detection~\citep{frintrop2015traditional,li2014secrets,wang2015deep}. The main advantage of saliency detection methods based on object proposals over methods based on superpixels \citep{yang2013saliency} is that they do not require an additional grouping phase, since the object proposals are expected to encompass the whole object. Other than general object detection, salient object segmentation aims at detecting objects which are salient in the scene.  Direct surrounding of objects is of importance to determine the object's saliency. Therefore, in this paper we extend the usage of object proposals for saliency detection with context proposals, which allow us to directly assess the saliency of the object with respect to its context.
\section{Method Overview}\label{sec:overview}

\begin{figure}[t] 
\centering
\includegraphics[width=0.92\textwidth ]{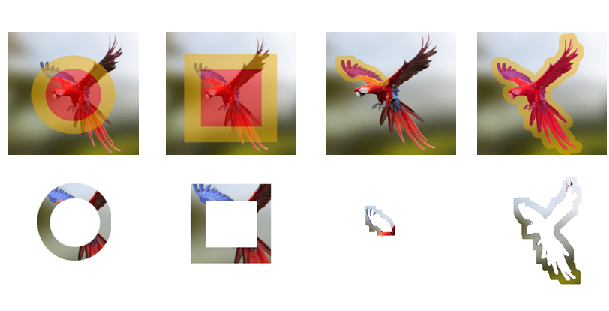}
\caption{Top row: examples of different center surround approaches (a) circular surround (b) rectangular surround (c) superpixels surround, and (d) the context proposals. Bottom row: the surround for each of the methods. It can be seen that only the object proposal based surround correctly separates object from background.}
\label{fig:context2}
\end{figure}

The main novelty of our method is the computation of context features from  context proposals. To illustrate the advantage of this idea consider Fig.~\ref{fig:context2}. In this figure several implementation of the center surround idea for saliency detection are shown. The circular surround was used in the original work by~\cite{itti1998model}. This concept was later generalized to arbitrary rectangles~\citep{liu2011learning}. Both these approaches have the drawback that they only are a rough approximation of the real object shape and the contrast between the (circle or rectangular) object and its surround does not very well represent the real saliency of the object. This is caused by the fact that when we approximate the object by either a square or circle, part of the object is in the surround, and part of the surround is in the object. 

In principle the center surround idea could be extended to superpixels which are often used in saliency detection~\citep{yang2013saliency}, see Fig.~\ref{fig:context2}. However, superpixels generally only cover a part of the object, and therefore their surround is often not homogeneous, complicating the analysis of the saliency of the center. Finally,~\cite{mairon2014closer} show that a surround which can adapt to the shape of the object (center) is an excellent saliency predictor. For its computation they propose an iterative procedure. In this paper we propose to use object proposals methods~\citep{arbelaez2014multiscale}, which are designed to directly provide a segmentation of the object, for the computation of context-based saliency. Since object proposals have the potential to correctly separate object from surround (see final column on the right in Fig.~\ref{fig:context2}), we hypothesize that considering their contrast can lead to a better saliency assessment than with the other methods. 

\begin{figure}[t]
\centering
\includegraphics[width=1\textwidth ]{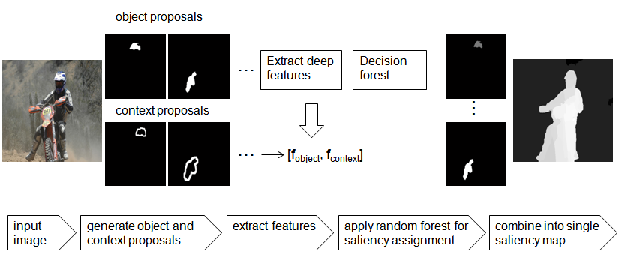}

\caption{Overview of our method at test time. A set of object proposals is computed. From these a set of accompanying context proposals is derived. We extract deep convolutional features from both object and context (${\bf f}_{object}$ and ${\bf f}_{context}$). At training for each object proposal its saliency is computed based on the ground truth, and a random forest is trained to regress to the saliency. At testing this random forest is applied to predict the saliency of all proposals, which are combined in the final saliency map}
\label{fig:framework}
\end{figure}

An overview of the saliency detection algorithm is provided in Fig.~\ref{fig:framework}. Next, any object proposal algorithm can be used here that provides pixel-precise object contours, such as \citep{arbelaez2014multiscale,uijlings2013selective,Krahenbuhl2014Geodesic,hu2016fastmask,pinheiro2016learning}. We extend each object proposal with a context proposal which is its immediate surround (see Section~\ref{sec:context_proposals}). We then proceed by computing deep features for both the object proposal and its context proposal from which we derive several context features (see Section~\ref{sec:context_features}). 

Given the feature vector of the object and context for each of the proposals in the training set we train a random forest classifier. As the saliency score for each object proposal we use the average saliency of the pixels in the proposal: pixels have a saliency of one if they are on the ground truth salient object or zero elsewhere (this procedure is further explained in Section~\ref{sec:saliency_score}). At testing time we infer the saliency for all the object proposals by applying the random forest regressor. The final saliency map is computed by taking for each pixel the average of the saliency of all the proposals that contain that pixel. 

The overall method is similar to several previous papers on saliency. A similar approach was proposed by~\cite{Jiang2013salient} and later used by ~\cite{li2014secrets}. In~\citep{Jiang2013salient} they use a random forest classifier to score each region in the image instead of every object proposal in our method. ~\cite{li2014secrets} use the CPMC method for object proposals~\citep{carreira2010constrained} and similar as~\cite{Jiang2013salient} they apply a random forest to predict region saliency based on regional features. In contrast to these methods we investigate the usage of context proposal for object saliency detection.

\section{Context Proposals for Saliency Computation}\label{sec:context}
In this section we start by describing our approach for context proposal generation. Then in Section~\ref{sec:context_features} we describe how to compute the context features from the context proposals. Next in Section~\ref{sec:shelf} we describe the deep features we directly use as features for the object proposals, and which we also use to compute the context features. Finally, in Section~\ref{sec:saliency_score} we explain how we arrive  at the final saliency estimation by using a random forest regressor on both the object and context features.

\subsection{Context Proposal Generation}\label{sec:context_proposals}

Recently, several saliency methods have applied object proposal algorithms to generate proposals for salient objects \citep{li2014secrets,Jiang2013salient}. Consider an object proposal, represented by the mask $M$ which is equal to one for all pixels within the object proposal and zero otherwise. Then we define the context of the proposal to be 
\begin{equation}
\begin{array}{l}
 C = \left( {M \oplus B^{(n)}} \right)\backslash M \\ 
 {\rm{smallest\;}}n{\;\rm{ for\;which\;}}\left| C \right| \ge \left| M \right| \\ 
 \end{array}
\end{equation}
where $B$ is a structural element and $\oplus$ is the dilation operator. We used the notation
\begin{equation}
B^{(n)}  = \overbrace{B^{(1)}\oplus B ^{(1)} \oplus B^{(1)} \oplus B^{(1)}}^{n\;times}
\end{equation}
to indicate multiple dilations. In our work we choose $B=N_8$ which is the eight connected set (a 3x3 structural element with all ones). We use $\left| C \right|$ to indicate the number of non-zero values in $C$. If we would consider arbitrary $n$ in the first part of this equation, this equation could be interpreted as generating a border for the object proposal $M$ which thickness is equal to $n$. We define the context to be the smallest border which has equal or more pixels than $M$. In practice, the context is computed by iteratively dilating with $B$ until we reach a border which contains more pixels than the object proposal $M$.
\begin{figure}[t] 
\centering
\includegraphics[width=1\textwidth ]{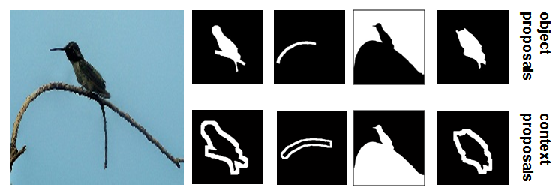} 
\caption{Input image and (top row) examples of object proposals and (bottom row) examples of context proposals.}
\label{fig:object_proposals}
\end{figure}
In Fig.~\ref{fig:object_proposals} we provide examples of context borders for several object proposals. Note that the context border is wider for larger object proposals. The idea is to verify if the object proposal is salient with respect to its context.

\subsection{Context Feature Computation}\label{sec:context_features}
Next we outline the computation of the context features. We consider two properties which define a good context proposal. {\it Context contrast} which measures the contrast between the features which make up the salient object and the features which describe its context. Secondly {\it Context continuity} which is based on the observation that the salient object is often occluding a background which continues behind it. As a consequence, we expect the features which describe the context on opposite sides of the salient object to be similar. In human vision research it was verified that salient objects (targets) are faster found on a homogeneous background than when surrounded by a heterogeneous background (distractor)~\citep{duncan1989visual}. Context continuity is an indicator of background homogeneity, since homogeneous backgrounds lead to higher context continuity, and heterogeneous ones would lead to low context continuity.

\begin{figure}[t]
\centering
\includegraphics[width=0.34\textwidth]{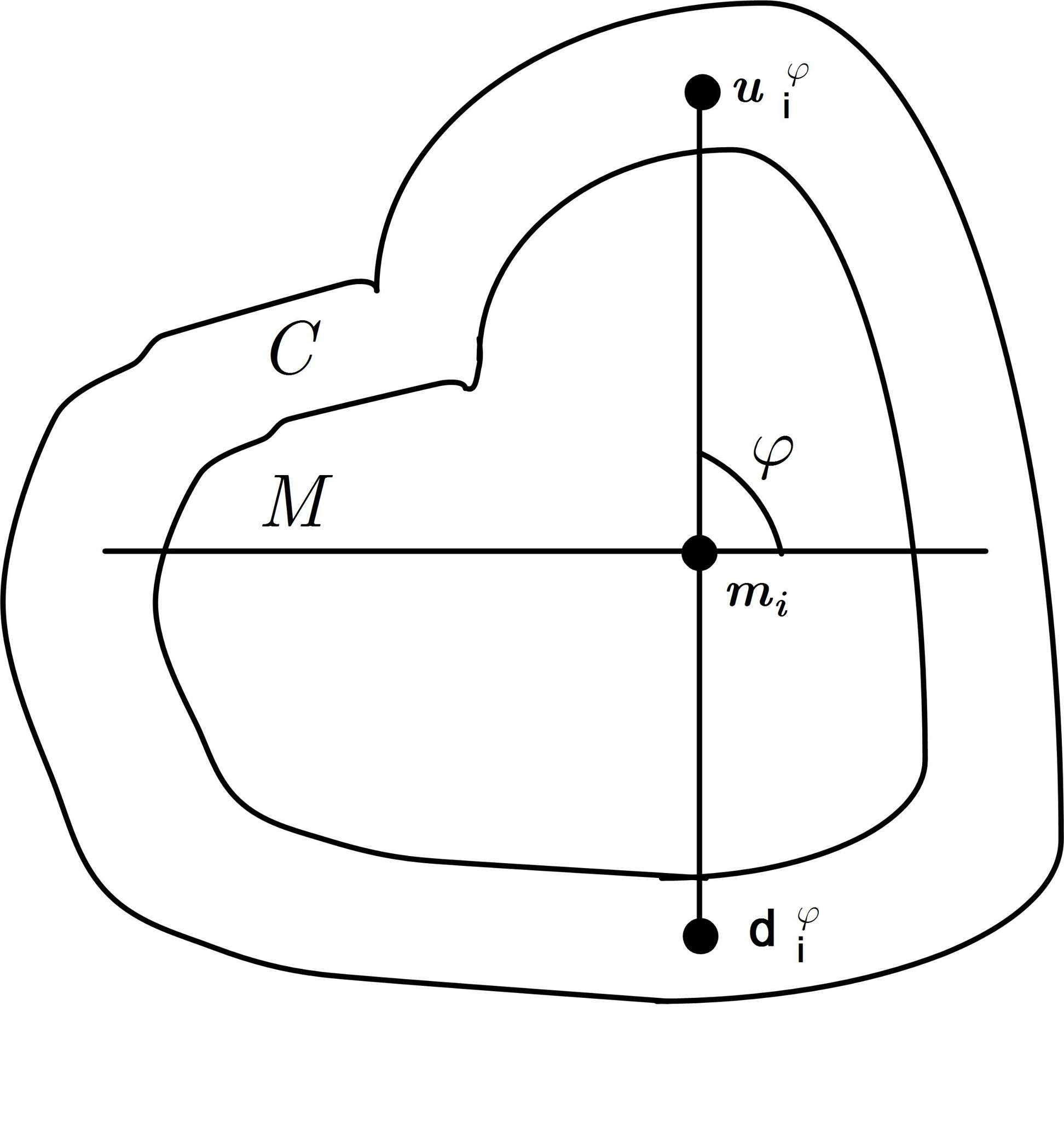}
\caption{Graphical representation of variables involved in context feature computation.}
\label{fig:context_feature}
\end{figure}

The first context saliency feature which we consider combines both described properties, context contrast and context continuity, into a single measure. Consider a pixel $m_i$ in the object proposal $M$. Then we define two related coordinates $d_i^{\varphi}$ and $u_i^{\varphi}$ which are coordinates of the points on the context when considering a line with orientation $\varphi$ through point $m_i$ (see Fig.~\ref{fig:context_feature}). The saliency of a point $m_i$ is larger when the feature representation at $m_i$ is more different from the feature representation on its context at $d_i$ and $u_i$. In addition, we would like the distance between the points on the context ($d_i$ and $u_i$) to be similar. Combining these two factors in one saliency measures yields:

\begin{equation}
c_1^\varphi \left( {m_i } \right) = \arctan \left(
\frac{  {\min \left( {s_i^{d,\varphi } ,s_i^{u,\varphi } } \right)} }{{s_i^{du,\varphi }  + \lambda }} \right).\label{eq:context1}
\end{equation}
where the numerator contains the context contrast and the denominator the context continuity.  The $arctan$ and the constant $\lambda$ are used to prevent large fluctuations in saliency for small values of $s_i^{du,\varphi }$. The distances are defined with
\begin{equation}
 s_i^{u,\varphi }  = \left\| {{\bf{f}}\left( {u_i^{\varphi } } \right) - {\bf{f}}\left( {m_i } \right)} \right\|, \\
 \label{eq:dist1}
 \end{equation}

 \begin{equation}
 s_i^{d,\varphi }  = \left\| {{\bf{f}}\left( {d_i^{\varphi } } \right) - {\bf{f}}\left( {m_i } \right)} \right\|, \\
 \label{eq:dist2}
 \end{equation}

 \begin{equation}
 s_i^{du,\varphi }  = \left\| {{\bf{f}}\left( {d_i^{\varphi } } \right) - {\bf{f}}\left( {u_i^{\varphi } } \right)} \right\| .
 \label{eq:dist3}
\end{equation}

Here  ${\bf{f}}\left(  {m_i } \right)$ denotes a feature representation of the image at spatial location $m_i$, and $\left\| {.} \right\|$ is the L2 norm. This feature representation could for example be the RGB value at that spatial location, but also any other feature representation such as for example a deep convolutional feature representation as we will use in this article. 
Now that we have defined the saliency for a single point considering its context points along a line with orientation $\varphi$, we define the overall saliency for a context proposal as the summation over all pixels $m_i$ in the object proposal considering all possible lines: 
\begin{equation}
C^1 = \frac{1}{\left| M \right|} \sum\limits_{m_i  \in M} {\int\limits_0^\pi  {c_1^\varphi  \left( {m_i } \right)d\varphi } } .
\end{equation}
It should be noted that we exclude lines which do not have context on both sides of the object. This happens for example for objects on the border of the image. 

Considering all orientations is computationally unrealistic and in practice we approximate this equation with 
\begin{equation}
C^1  = \frac{1}{\left| M \right|}\sum\limits_{m_i  \in M} {\sum\limits_{\varphi  \in \Phi } {c_1^\varphi  \left( {m_i } \right)} },
\label{eq:context-feature1}
\end{equation}
where $\Phi$ is a set of chosen orientations between $\left[ {0,\pi } \right)$. In this paper we have considered four orientations
\begin{equation}
\Phi  = \left\{ {0,\frac{\pi }{4},\frac{\pi }{2},\frac{{3\pi }}{4}} \right\}.
\end{equation}
The saliency of one point in the object proposal is hence computed by considering its context along four orientations. To be less sensitive to noise on the context both  ${\bf{f}}\left( {d_i^{\varphi } } \right)$ and $ {\bf{f}}\left( {u_i^{\varphi } } \right) $ are extracted from a Gaussian smoothed context proposal. 

As a second context feature we ignore the object proposal and only consider the values on the context proposal to compute the saliency. This feature solely focuses on context continuity. In this case we would like the saliency to be larger when the values on the context have a smaller distance. We propose to use the following measure:
\begin{equation}
c_2^\varphi  \left( {m_i } \right) = \arctan \left( \frac{1}{{s_i^{du,\varphi }  + \lambda }}\right)
\label{eq:context2}
\end{equation}
again $\lambda$  prevents large fluctuations for low values of $s_i$.

Similarly we compute the $C^2 \left( {m_i } \right)$ for the object proposal with
\begin{equation}
C^2  = \frac{1}{\left| M \right|} \sum\limits_{m_i  \in M} {\int\limits_0^\pi  {c_2^\varphi  \left( {m_i } \right)d\varphi } } .
\end{equation}
and its approximation
\begin{equation}
C^2  = \frac{1}{\left| M \right|} \sum\limits_{m_i  \in M} {\sum\limits_{\varphi  \in \Phi } {c_2^\varphi  \left( {m_i } \right)}. } 
\label{eq:context-feature2}
\end{equation}

\begin{figure}[tb]
\centering
\includegraphics[height=0.39\textwidth]{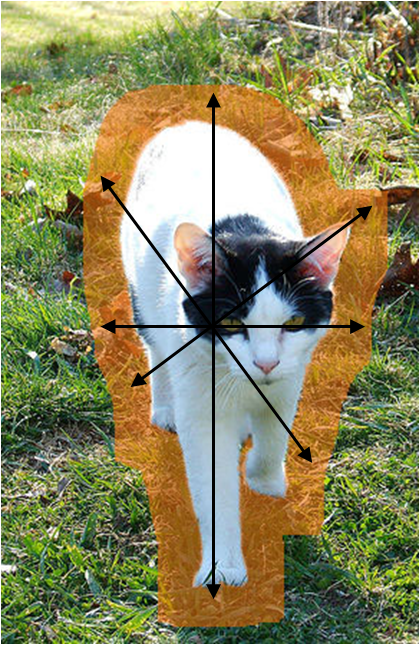}
\includegraphics[height=0.39\textwidth]{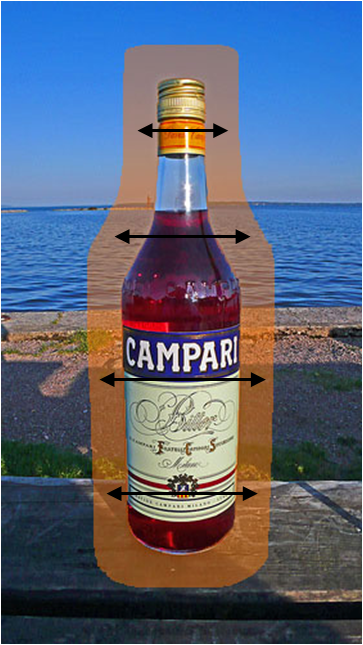}
\caption{Context continuity: features on opposites sides of the object proposal are expected to be similar. Examples of (left) omni-directional context continuity and (right) horizontal context continuity.}
\label{fig:horizontal_context}
\end{figure}

In addition to $C^1$ and $C^2$ which measure context saliency based on comparing features on all sides of the object proposal, we introduce also a measure for horizontal context continuity $C^3$ where we use $\Phi^H  = \left\{ 0 \right\}$, and we compute 
\begin{equation}
C^3  = \frac{1}{\left| M \right|}\sum\limits_{m_i  \in M} {\sum\limits_{\varphi  \in \Phi^H } {c_1^\varphi  \left( {m_i } \right)} }.
\label{eq:context-feature3}
\end{equation}
The motivation for a special measure for horizontal context continuity is provided in Fig.~\ref{fig:horizontal_context}. Natural scenes contain more horizontal elongation than other orientations; the $C^3$ measure is designed to detect horizontal clutter.

The context measures proposed here are motivated by the work of~\cite{mairon2014closer}. They propose an iterative procedure to compute context based saliency. We prevent the iterative procedure by directly computing the context from the object proposals. In addition, we propose a measure of horizontal context which is not present in~\citep{mairon2014closer}. Also instead of RGB features we use deep features to compute the context saliency.

\begin{table}[t]
\small
    \centering
    \begin{tabular}{|l|c|c|c|}\hline
\diagbox[width=3.2em]{bl.}{net.} &   AlexNet & VGG-19 & ResNet-152 \\\hline

1.& $ \left[11\mkern-2mu\times\mkern-2mu11,\mkern-2mu 96\right]$   & $\left[3\times3, 64\right]\mkern-2mu\times\mkern-2mu 2$ & $\left[7\times7, 64\right]\mkern-8mu$ \\ 
\hline

2.&$\left[5\times5, 256\right]$ & $\left[3\times3, 128 \right]\mkern-2mu\times\mkern-2mu2$ &  $\left[\begin{array}{lll} 1 \times 1, 64 \\ 3\times3, 64 \\1\times1,256\end{array} \right]\mkern-2mu\times\mkern-2mu\;3\mkern-8mu$ \\
\hline

3.& $\left[3\times3, 384\right]$       &  $\left[3\times3, 256 \right]\mkern-2mu\times\mkern-2mu4$ & $\left[\begin{array}{ccc} 1\times1, 128 \\ 3\times3, 128 \\1\times1,512\end{array} \right]\mkern-2mu\times\mkern-2mu\;8\mkern-8mu$\\ 
\hline

4.& $\left[3\times3, 384\right]$ & $\left[3\times3, 512 \right]\mkern-2mu\times\mkern-2mu4$ &  $\left[\begin{array}{ccc} 1\times1, 256 \\ 3\times3, 256 \\1\times1,1024\end{array}\right]\mkern-2mu\times\mkern-2mu 36\mkern-8mu$ \\
\hline
5.& $\left[3\times3, 256\right]$ & $\left[3\times3, 512 \right]\mkern-2mu\times\mkern-2mu4$ & $\left[\begin{array}{ccc} 1\times1, 512 \\ 3\times3, 512 \\1\times1,2048\end{array}\right]\mkern-2mu\times\mkern-2mu \;3\mkern-8mu$\\
\hline

\end{tabular}
    \caption{Overview of the \emph{convolutional} layers of different networks. The convolutional part can be divided in 5 blocks (bl.) for all three networks. For each block we show the convolutional size, the number of features, and how many times this layer pattern is repeated. The non-linear activation layers are omitted. In our evaluation we will use the last layer of each block to extract convolutional features. }
    \label{tab:overview}
\end{table}

\subsection{Off-the-Shelf Deep Features}\label{sec:shelf}
Here we explain the computation of the deep features, which we use as the feature ${\bf f}$ in Eq.~\ref{eq:dist1}-\ref{eq:dist3} to compute the three context features Eq.~\ref{eq:context-feature1}, Eq.~\ref{eq:context-feature2}-~\ref{eq:context-feature3}. These are combined into one context feature
\begin{equation}
{\bf f}_{context}=\{C^1,C^2,C^3\}\label{eq:f_context}
\end{equation}
for each context proposal. The deep feature is also used directly as a descriptor for the object proposal by pooling the deep feature over all pixels in the object proposal with
\begin{equation}
{\bf f}_{object}= \frac{1}{\left| M \right|}\sum\limits_{m_i  \in M} { {\bf f}  \left( {m_i } \right)}. \label{eq:f_object}.
\end{equation}
Deep convolutional features have shown excellent results in recent papers on saliency  ~\citep{wang2015deep,li2015visual,zhao2015saliency,LiYu2016}. A straight-forward way to use deep features is by using a pre-trained network, for example trained for the task of image classification on ImageNet~\citep{KrizhevskySutskever2012}, to extract features. These so called off-the-shelf features can then be used as local features. A good overview of this approach is given by ~\cite{SharifAzizpour2014}, who successfully apply this technique to a variety of tasks including object image classification, scene recognition, fine grained recognition, attribute detection and image retrieval. 
 
\begin{figure}[t]
\centering
\includegraphics[width=.9\textwidth]{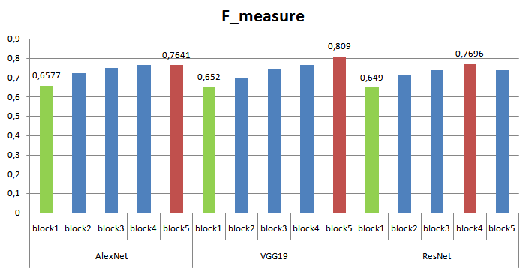}
\caption{Evaluation on 5 convolutional layers for the three architectures used in our framework.}
\label{fig:Eval}
\end{figure}

To choose the best deep features for saliency detection we evaluate three popular networks, namely AlexNet~\citep{KrizhevskySutskever2012}, VGG-19~\citep{SimonyanZisserman2014} and ResNet~\citep{he2016deep}. The configuration of the convolutional layers of the networks is given in Table~\ref{tab:overview}. We evaluate the performance of the different blocks for saliency estimation. The results using both object features ${\bf f}_{object}$ and context features ${\bf f}_{context}$ are summarized in Fig.~\ref{fig:Eval}. We found the best results, similar to the ResNet, were obtained with block 5 of VGG-19 (which layer name is \textit{conv5\_4}). Based on these results we choose to extract block 5 deep features with VGG-19 for all images. We spatially pool the features within each object to form a 512-dimensional ${\bf f}_{object}$ and the 3-dimensional  ${\bf f}_{context}$ according to Eq.~\ref{eq:f_context}-~\ref{eq:f_object}. In addition, we found that applying a standard whitening, where we set the variance over all features of ${\bf f}_{object}$ to 1, prior to applying the classifiers improved results.

\subsection{Saliency Score of Object Proposals}\label{sec:saliency_score}
Based on the features which are extracted from the object proposal and its
context we train a random forest regressor to estimate the saliency 
of the object proposal. To compute the saliency score, $sal^{object}$, for object proposals we use
the following equation:
\begin{equation}
sal^{object} = \frac{{\left| {M \cap S} \right|}}{{\left| M \right|}}
\label{Eq:eval1}
\end{equation}
here $M$ is the set of all pixels in the object proposal and $S$ is the set 
of all pixels which are considered salient in the ground truth. A $sal=0.8$
means that 80\% of the pixels in the object proposal are considered salient.

We found that this score is not optimal when considering context proposals, and we propose to use the following equation
\begin{equation}
sal^{context}  = \max \left( {\frac{{\left| {M \cap S} \right|}}{{\left| M \right|}} - \frac{{\left| {C \cap S} \right|}}{{\left| C \right|}},0} \right)
\label{Eq:eval2}
\end{equation}
where $C$ is the set of pixels in the context. The $sal^{context}$ measure
lowers the score if salient pixels are in the context. 

We train two separate random forest regressors, one based on the deep features of the 
object proposal regressing to $sal^{object}$ and one based on the context features regressing to 
$sal^{context}$. The final saliency score at testing time is computed by adding results of the two regressors.
The final saliency map is computed by averaging the saliency of all the object proposals which are considered in the image. We have also considered to assign to each pixel the maximum saliency of all object proposals which include the pixel, but found this to yield inferior results.


\section{Experimental Setup}\label{sec:Exp_setup}
In this section we describe the features on which we base the saliency computation, the datasets on which the experiments are performed, and the evaluation protocol we use.

\subsection{Datasets}
We evaluate our proposed algorithm on several benchmark datasets that are widely used.

\minisection{Pascal-S~\citep{li2014secrets}:}This dataset was built on the validation set of the Pascal VOC 2010 segmentation challenge. It contains 850 images with both saliency segmentation ground truth and eye fixation ground truth. Saliency ground truth masks were labeled by 12 subjects. Many of the images in this dataset contain multiple salient objects. 
 
\minisection{MSRA-B~\citep{liu2011learning}:}This dataset contains 5,000 images and is one of the most used datasets for visual saliency estimation. Most of the images contain only one salient object.

\minisection{FT~\citep{Achanta2009}:}This dataset contains 1,000 images, most of the images contain one salient object. It provides only salient object ground truth which is derived from \cite{wang2008two} and is obtained using user-drawn rectangles around salient objects.

\minisection{ECSSD~\citep{yan2013hierarchical}:}It contains 1,000 images acquired from the PASCAL VOC dataset and the internet and the ground truth masks were annotated by 5 subjects.

\subsection{Evaluation}
We evaluate the performance using PR (precision-recall) curve and F-measure.  Precision measures the percentage of salient pixels correctly assigned, and recall the section of detected salient pixels which belongs to the salient object in the ground truth.

We compute precision and recall of saliency maps by segmenting the salient object with a threshold $T$  and comparing the binary map with the ground truth. All saliency maps are also evaluated using the F-measure score which is defined as:
\begin{equation}
F_{\beta} \: = \: \frac{\left( {1+\beta^2 } \right) \cdot \mathrm{precision} \cdot \mathrm{recall}} {\beta^2 \cdot \mathrm{precision} + \mathrm{recall} }
\end{equation}
where $\beta^2$  is set to $0.3$ following \citep{li2014secrets,wang2015deep, sun2015saliency,zhao2015saliency}. As a threshold we use the one which leads to the best $F_{\beta}$. This was proposed in \citep{borji2012salient,martin2004learning} as a good summary of the precision-recall curve. We compare our method against 8 recent CNN methods: Deeply supervised salient object (DSS)\citep{hou2017deeply}, Deep contrast learning (DCL) \citep{Li2016deep}, Reccurent fully convolutional networks (RFCN) \citep{wang2016saliency}, Deep hierarchical salieny (DHS) \citep{liu2016dhsnet}, Multi-task deep saliency (MTDS) \citep{li2016deepsaliency}, Multiscale deep features (MDF) \citep{li2015visual}, Local and global estimation (LEGS) \citep{wang2015deep}, Multi context (MC) \citep{zhao2015saliency} and we compare also against 8 classical methods including Discriminative regional feature integration (DRFI) \citep{Jiang2013salient}, Hierarchical saliency (HS) \citep{yan2013hierarchical}, Frequency tuned saliency (FT) \citep{Achanta2009}, Regional principal color based saliency detection (RPC) \citep{lou2014regional}, (CPMC-GBVS) \citep{li2014secrets}, Graph-based manifold ranking (GBMR) \citep{yang2013saliency}, Principal component analysis saliency (PCAS) \citep{margolin2013makes}, Textural distinctiveness (TD) \citep{scharfenberger2013statistical} and a Context aware method \citep{goferman2012context} (GOF). For a fair comparison we did not include (CPMC-GBVS) method\citep{li2014secrets}  because they use eye fixation label in training.

Based on crossvalidation experiments on PASCAL-S training set we set the number of trees in the random forest to 200, we set $\lambda=40$ in Eq.~\ref{eq:context1} and Eq.~\ref{eq:context2} and we set the minimum area of object proposals to be considered at 4,500 pixels. We use these settings for all datasets.


\section{Experimental Results}\label{sec:Exp}
In this section we provide our experimental results. We provide an evaluation of five popular object proposal approaches. Next we evaluate the relative gain which is obtained by adding the features based on context proposals. We evaluate also our context features with different context shapes including the conventional circular or rectangular neighborhood. Finally, we compare to state-of-the-art methods on several benchmark datasets.

\subsection{Object Proposal based Saliency Detection}\label{sec:exp_proposals}
\minisection{Object proposal method evaluation:}In recent years several methods have proposed to use object proposals for salient object segmentation. However, to the best of our knowledge, there is no work which evaluates the different object proposal approaches to saliency detection.~\cite{hosang2015makes} have provided an extensive evaluation of object proposals for object detection. Based on their analysis we have selected the three best object proposal methods which output segments based on their criteria, namely repeatability, recall, and detection results. The object proposal methods we compare to are selective search (SS)~\citep{uijlings2013selective}, the geodesic object proposals (GOP)~\citep{Krahenbuhl2014Geodesic}, and the multiscale combinatorial grouping (MCG) method~\citep{arbelaez2014multiscale}. We have added two recent object proposals to this list which are based on deep learning, namely FastMask \citep{hu2016fastmask} and SharpMask \citep{pinheiro2016learning}. We do these experiments on the PASCAL-S dataset because it is considered one of the most challenging saliency datasets; also it is labeled by multiple subjects without restriction on the number of salient objects~\citep{li2014secrets}.

We evaluate the performance of object proposal methods as a function of proposals. Results are provided in Table.~\ref{tab:Object_proposals2}. Results of MCG are remarkable already for as few as 16 proposals per image, and they stay above the other methods when increasing the number of proposals. The results of SS can be explained by the fact that the ranking of their proposals is inferior to the other methods. The inferior ranking is not that relevant for object detection where typically thousands of proposals are considered per image\footnote{Selective search applies a pseudo random sorting which combines random ranking with a ranking based on the hierarchical merging process.}. The results of the two methods based on deep learning, namely FastMask and SharpMask, are somewhat surprising because they are known to obtain better results for object detection \citep{hu2016fastmask,pinheiro2016learning}. In a closer analysis we found that MCG obtains higher overlap (as defined by IoU) with the salient object groundtruth. In addition, deep learning approaches typically extract the salient object among the first 8-16 proposals, and therefore do not improve, and sometimes even deteriorate, when considering more proposals. Based on the results we select MCG to be applied on all further experiments, and we set the number of object proposals to 256.

\begin{table}[]
    \centering
    \begin{tabular}{|c|c|c|c|c|c|c|c|}\hline
    Number of proposals &  8 &  16& 32&64&128&256\\ \hline
     SS& 59.00& 64.60& 70.20& 74.20& 77.50& 78.40 \\ \hline
    GOP&66.20 & 71.50& 73.30& 76.30& 77.70 & 79.60\\ \hline
    MCG & 77.20& 77.50 &78.60& 79.30 &80.20&\bf{80.90}\\ \hline
    SharpMask & 73.79&74.07 &73.34&73.15&73.70&74.01\\ \hline
    FastMask &75.87 & 75.03&74.42&74.04& $-$&$-$\\ \hline

    \end{tabular}
    \caption{The F-measure performance as the number of proposals evaluated on the PASCAL-S dataset for selective search (SS), geodesic object proposals (GOP), multiscale combinatorial grouping (MCG), SharpMask and FastMask }
    \label{tab:Object_proposals2}
\end{table}

\begin{figure}[!htbp]
\centering
\includegraphics[width=0.75\textwidth]{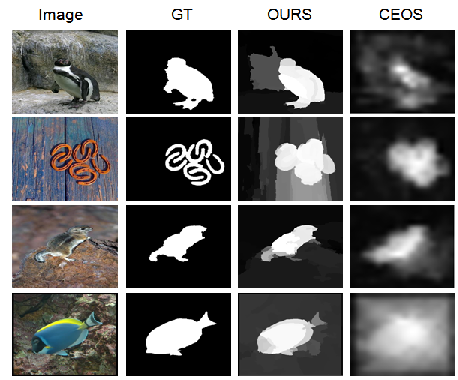}
\caption{Visual comparison between our method and the method of \cite{mairon2014closer}. Our method results in clearer edges since saliency is assigned to whole object proposals.}
\label{fig:context comparison}
\end{figure}

\minisection{Context proposals:} 
The proposed context features are motivated by the work of \cite{mairon2014closer}. Different from it, our paper does not use an iterative procedure but is based on object proposals. We add a comparison in Table~\ref{tab:context comp} of the performance of our context features against their method on the PASCAL-S dataset. Note that here we only consider our context feature for a fair comparison, and do not use the object feature.  We have also included results when only using RGB features, which are the features used by \citep{mairon2014closer}. Our context features clearly outperform the context features based on both RGB and deep features. We have also included timings of our algorithm. Since most of the time was spend by the MCG algorithm (35.3s) we have also included results with the FastMask object proposals (using 8 proposals). In this case the computation of the context features takes (5.4s). Note that this is based on an unoptimized matlab implementation. Also we add a visual comparison between our method and \citep{mairon2014closer} in Fig.~\ref{fig:context comparison}. 

Next we compare our context proposals, which follow the object proposal boundary, with different context shapes. We consider rectangular and circular context, which are derived from the bounding boxes based on the object proposals \footnote{The context of the rectangular bounding box is computed by considering its difference with a rectangle which is $\sqrt 2 $ larger. In case of the circular context we consider the circle center to have a radius of $r=\frac{w+h}{4}$ and its context is computed by considering the difference with a radius larger by a factor of $\sqrt 2 $. Like this the context for both the rectangle and the circle has again the same surface area as the object (center).}. For the three different context shapes we extract the same context features. The results are summarized in Table~\ref{tab:context circular-rectangular} and show that our approach clearly outperforms the rectangular and circular shaped contexts. Thereby showing that accurate context masks result in more precise saliency estimations.

In the following experiment we evaluate the additional performance gain of the saliency features based on context proposals. The results are presented in Table~\ref{tab:relative_gain} for four datasets. We can see that a consistent performance gain is obtained by the usage of context proposals. The absolute gain varies from 0.7 
on FT to 1.6 on PASCAL-S. This is good considering that the context feature only has a dimensionality of 3 compared to 512 for the object feature.

\begin{table}
\centering
\begin{tabular}{|c|c|c|c|c|} \hline
& feature & proposals &PASCAL-S & Time(s) \\ \hline
Mairon &RGB&- &$65.57$ &140\\\hline
Our context & RGB & MCG & $69.06$ & 40.9\\ \hline
Our context & DF & MCG &  $74.90$  & 49.0\\ \hline
Our context & DF & FastMask & $73.65$ & 6.7 \\ \hline
\end{tabular}
\caption{Comparison between our context features and the context method proposed by \cite{mairon2014closer} in terms of F-measure and computational speed in seconds. We provide results for our method based on RGB and deep features (DF), and with MCG or FastMask as a object proposal method.}.
\label{tab:context comp}
\end{table}

\begin{table}
\centering
\begin{tabular}{|c|c|} \hline
Method & PASCAL-S   \\ \hline
Our context features  &  $74.90$ \\ \hline
Rectangular center surround & $67.64$ \\ \hline
Circular center surround & $63.71$ \\  \hline
\end{tabular}
\caption{Comparison between our context shape and the conventional circular or rectangular neighborhood in terms of F-measure.  }.
\label{tab:context circular-rectangular}
\end{table}

\begin{table}
\centering
\begin{tabular}{|c|c|c|c|} \hline
& object & context  &   object \&  context    \\ \hline 
PASCAL-S & $80.64$ & $74.90$ & $82.31$ \\ \hline
 MSRA-B & $89.90$ & $89.24$ & $90.90$ \\ \hline
 FT & $89.80$ & $87.96$ & $91.5$ \\ \hline
 ECSSD & $86$ & $82.64$ & $86.90$ \\ \hline
\end{tabular}
\caption{The results on four datasets in F-measure for saliency based only on object proposals, only context proposals and a combination of the two.}
\label{tab:relative_gain}
\end{table}

\begin{figure}[!htbp]
\centering
\includegraphics[width=.48\textwidth]{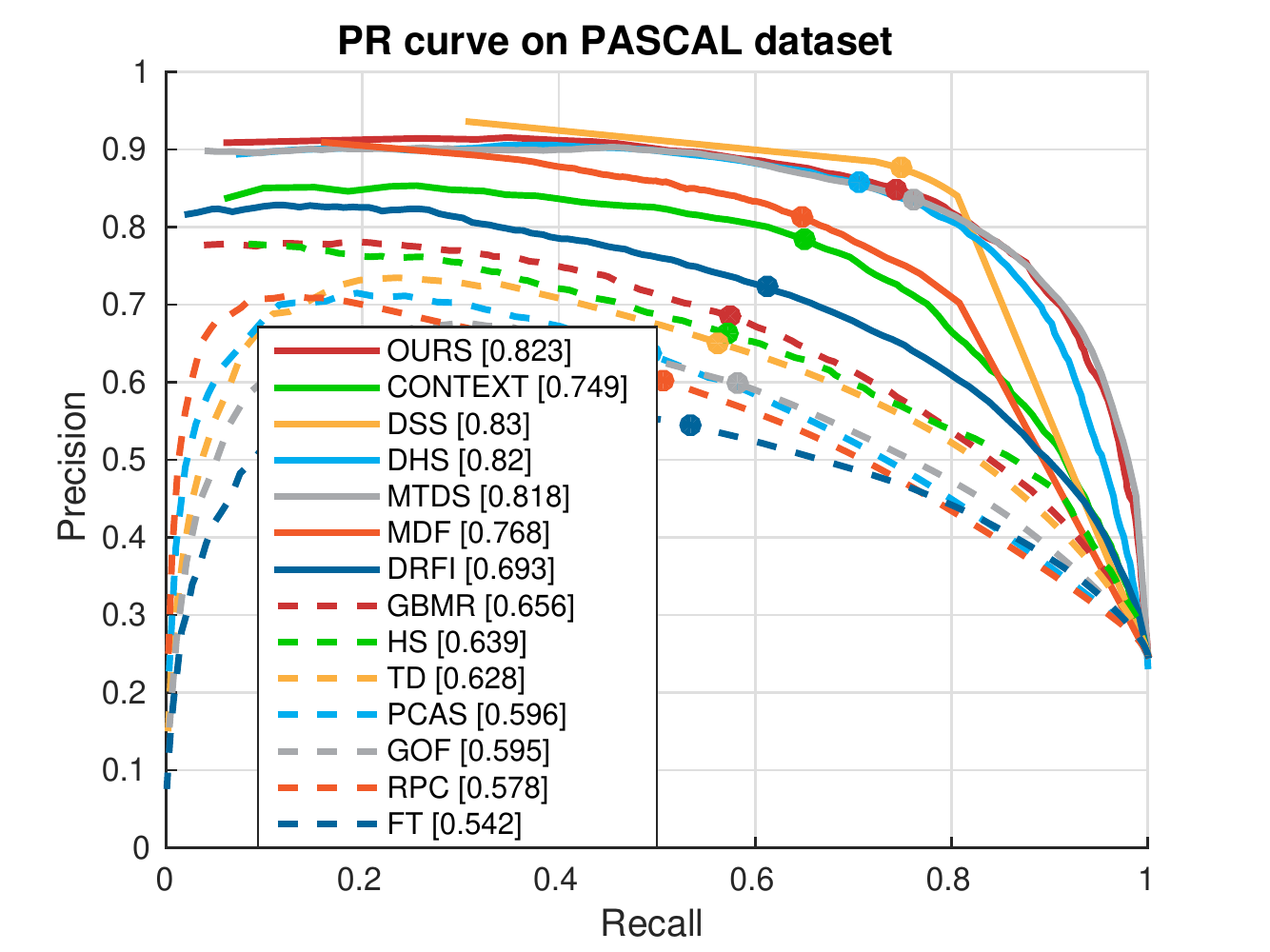}
\includegraphics[width=.48\textwidth ]{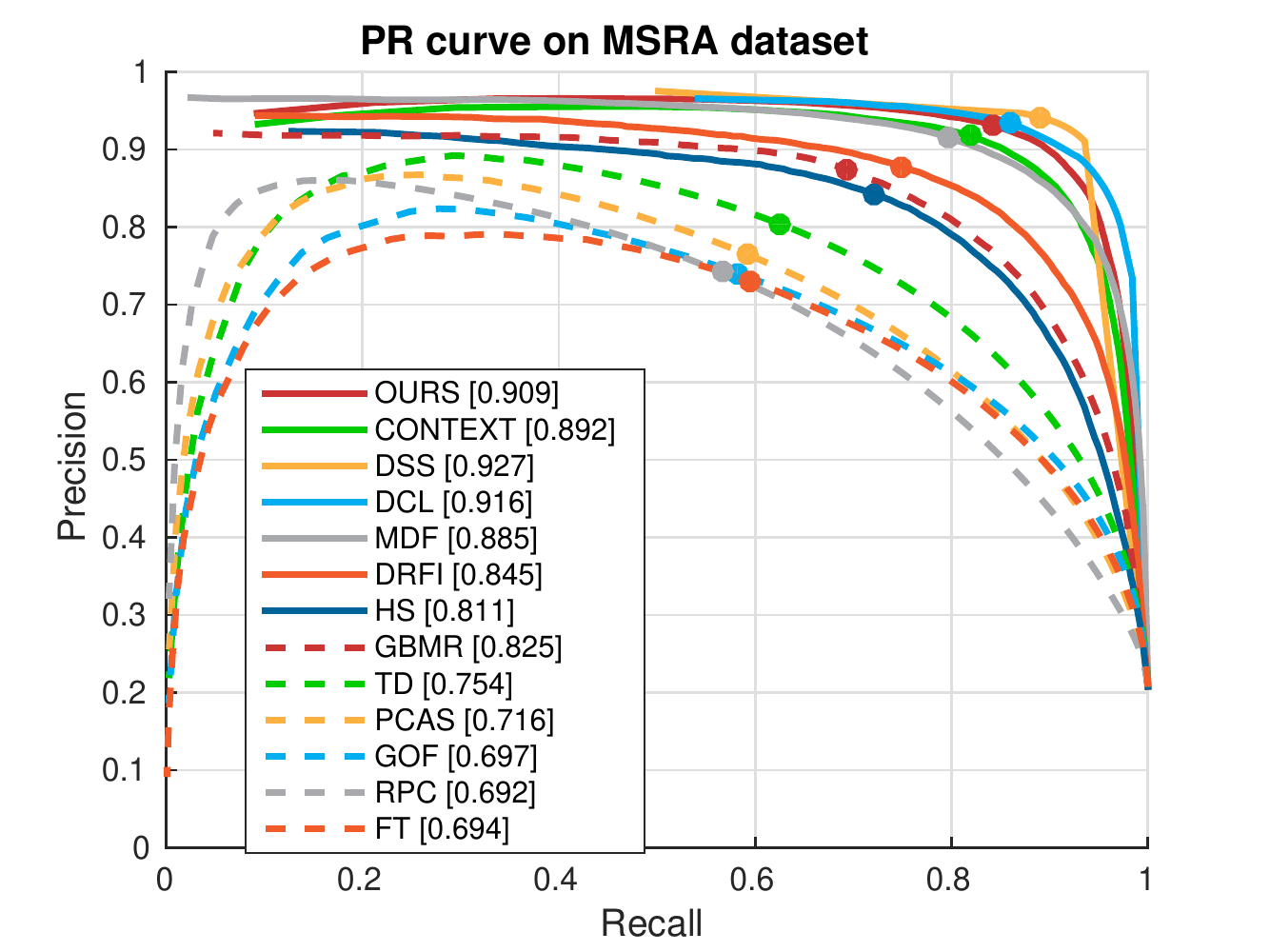}
\caption{Precision-Recall curves on (left) Pascal-S dataset and (right) on MSRA-B dataset}
\label{fig:pascal}
\end{figure}
\begin{figure}[!htbp]
\centering
\includegraphics[width=.48\textwidth]{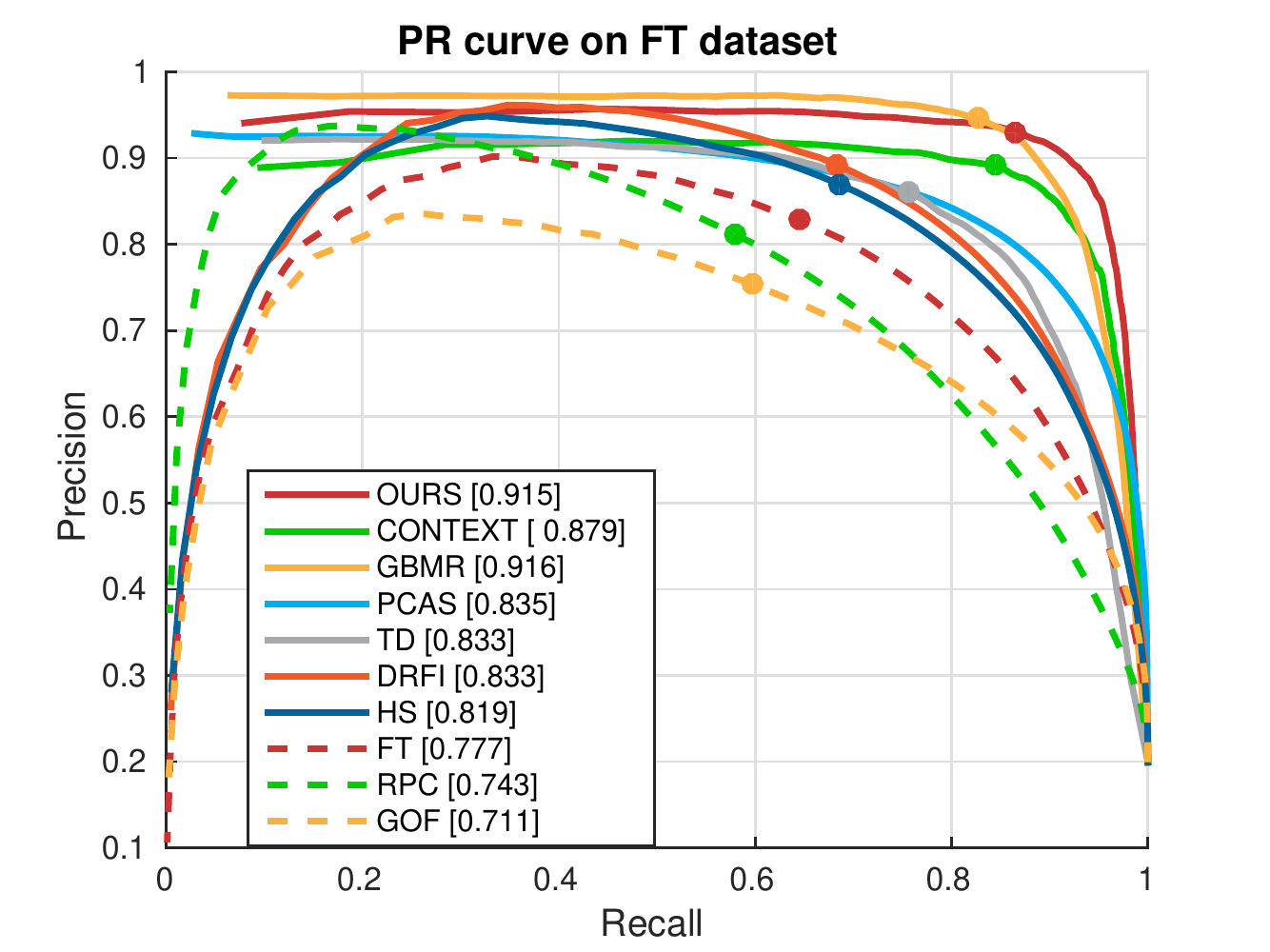}
\includegraphics[width=.48\textwidth]{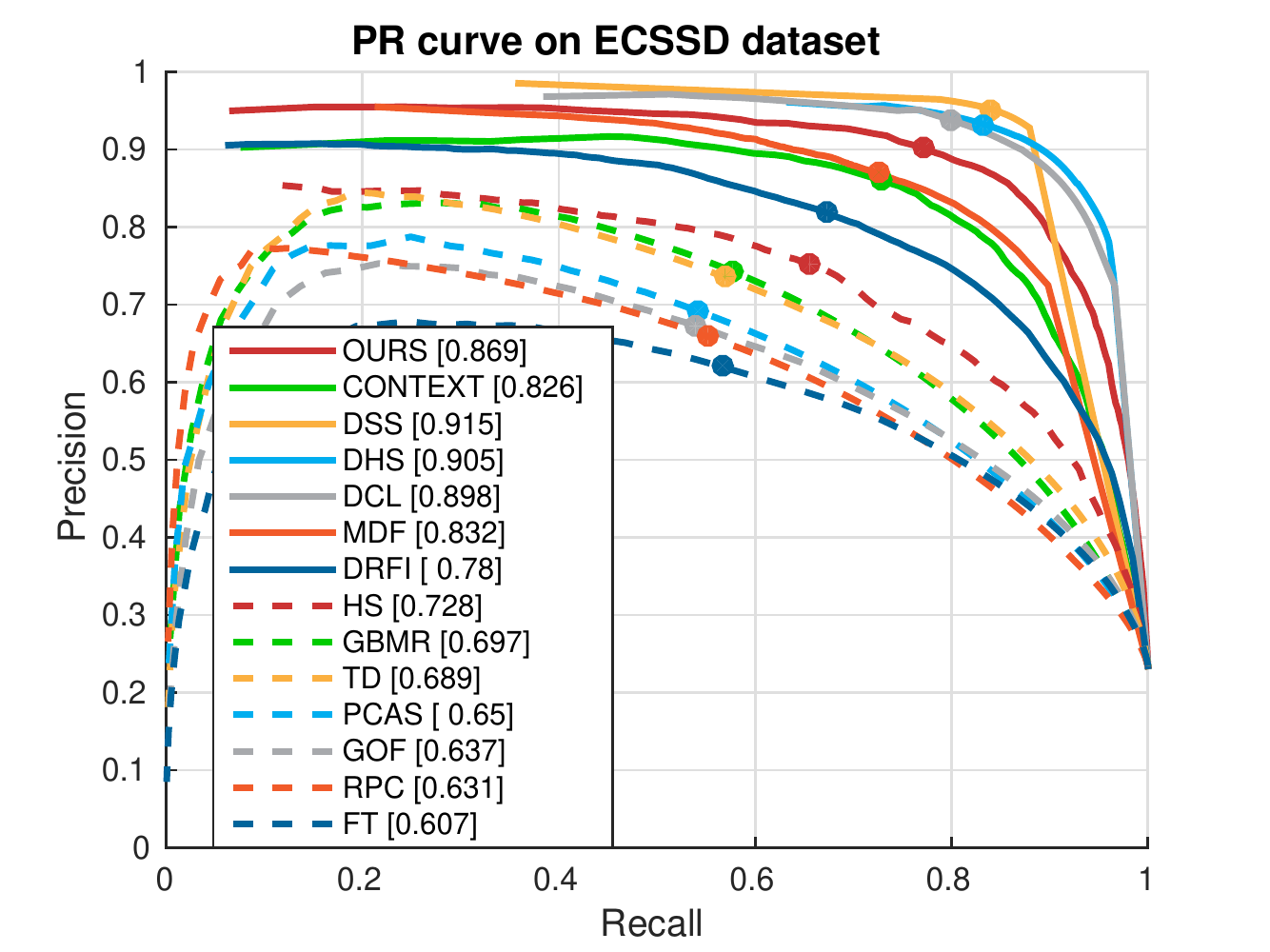}
\caption{Precision-Recall curves on (left) FT dataset and (right) ECSSD dataset.}
\label{fig:MSRA}
\end{figure}

\subsection{Comparison with the state-of-the-art}
Experiments have been conducted on the PASCAL-S, MSRA-B, FT and ECSSD datasets.  Traditionally these datasets proposed an original train and testset split~\citep{li2014secrets}. However, several of these datasets are too small to train deep neural networks. Therefore, methods based on deep learning generally train on the MSRA-B trainset which is the largest available dataset ~\citep{Jiang2013salient,li2015visual,Li2016deep}. To be able to compare with all results reported in the literature,  we report in Table~\ref{tab:comp} both results; the results trained on the original training set and those based on training on the MSRA-B training set (these results are indicated by an asterix). As an evaluation metric we use the F-measure. We report both qualitative and quantitative comparison of our methods with state-of-the-art methods. We also report our results in Figs.~\ref{fig:pascal}-\ref{fig:MSRA}. Note that these are based on training on the original training set of each dataset. Furthermore, we have only included the curves of the methods in Figs.~\ref{fig:pascal}-\ref{fig:MSRA} when this data is made available by the authors. 

On the challenging PASCAL-S dataset our method trained on the original dataset obtains an F-measure of 82.3, and is the third method. On the MSRA-B dataset we are outperformed by several recent end-to-end trained saliency methods but still obtain competitive results of $90.9$. On the FT dataset we obtain similar to state-of-the-art results when trained on the original dataset, and slightly better than state-of-the-art when trained on the MSRA-B dataset. Finally, on the ECSSD dataset we obtain the best results when considering only those which are trained on the ECSSD training dataset, but are outperformed by recent end-to-end trained networks trained on MSRA-B.

\begin{figure*}[!htbp]
\def\arraystretch{.75}
\setlength{\tabcolsep}{1mm}
\begin{tabularx}{\textwidth}{@{\extracolsep\fill}ccccccccccc}
\includegraphics[width=0.08\textwidth]{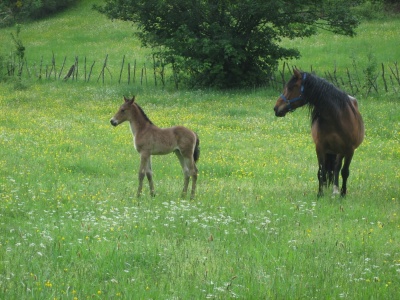} & \includegraphics [width=0.08\textwidth]{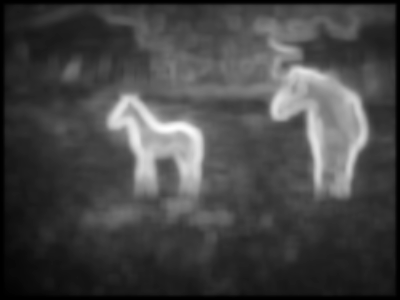} & \includegraphics [width=0.08\textwidth]{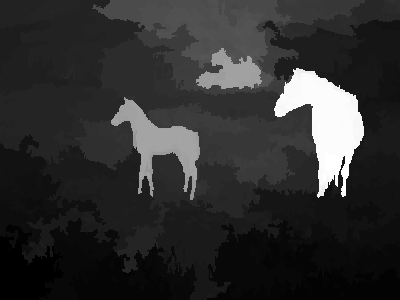} & \includegraphics [width=0.08\textwidth]{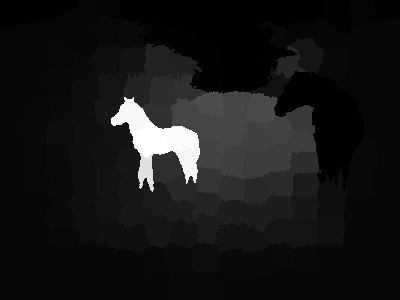}  &\includegraphics[width=0.08\textwidth]{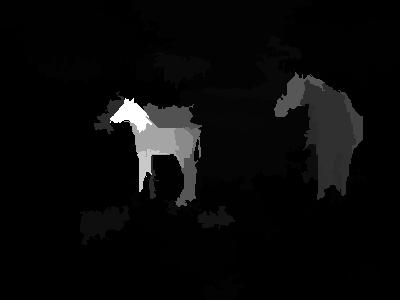} & \includegraphics [width=0.08\textwidth]{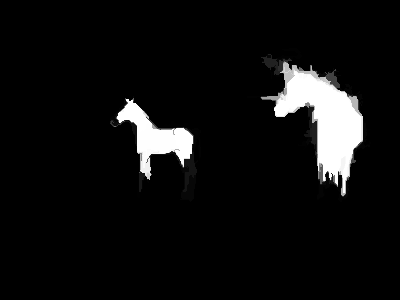} & \includegraphics [width=0.08\textwidth]{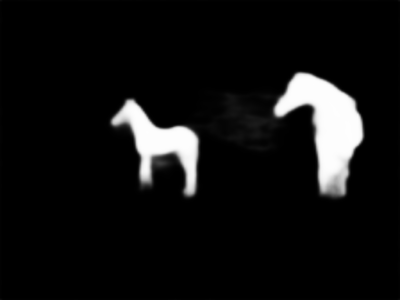} & \includegraphics [width=0.08\textwidth]{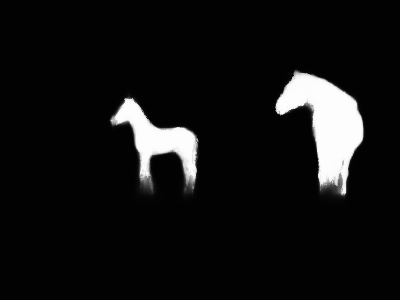} & \includegraphics[width=0.08\textwidth]{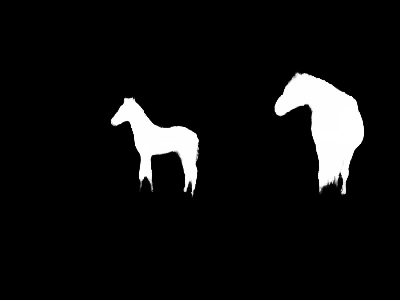} & \includegraphics [width=0.08\textwidth]{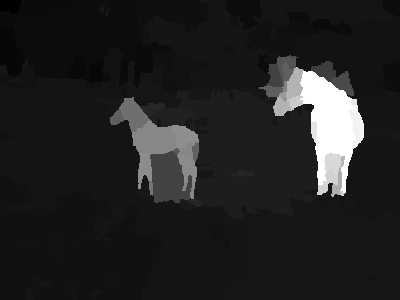} & \includegraphics [width=0.08\textwidth]{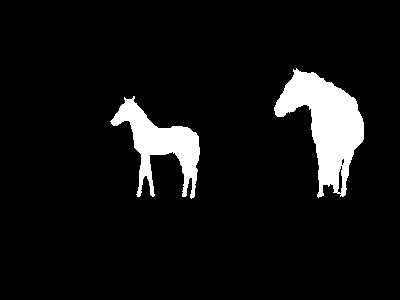} \\ 
\includegraphics[width=0.08\textwidth]{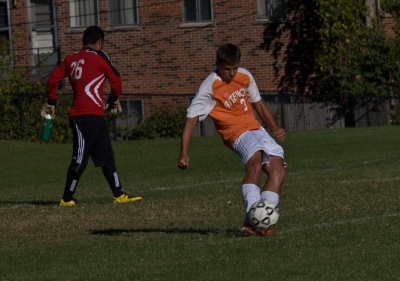} & \includegraphics [width=0.08\textwidth]{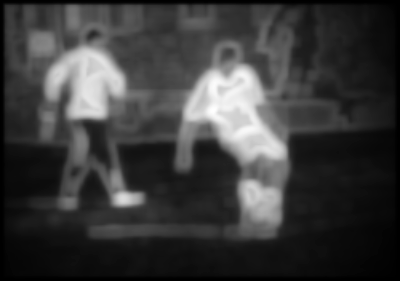} & \includegraphics [width=0.08\textwidth]{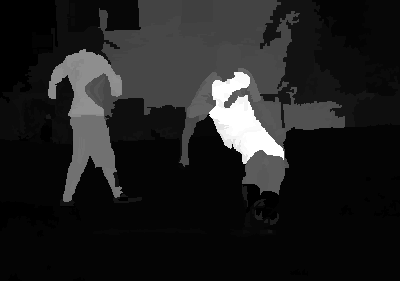} & \includegraphics [width=0.08\textwidth]{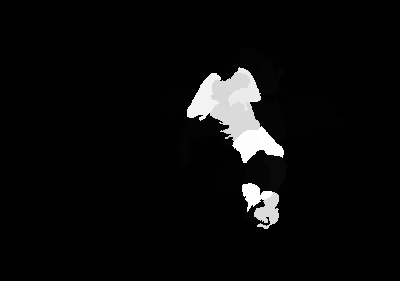}  &\includegraphics[width=0.08\textwidth]{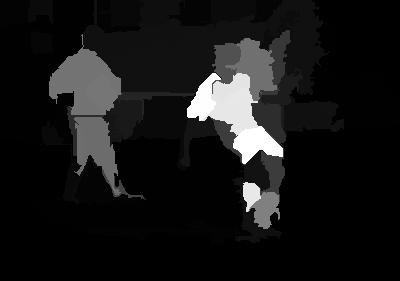} & \includegraphics [width=0.08\textwidth]{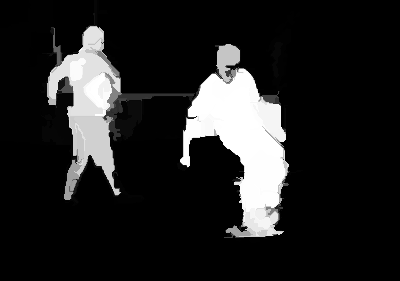} & \includegraphics [width=0.08\textwidth]{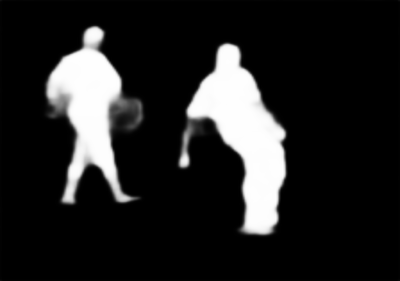} & \includegraphics [width=0.08\textwidth]{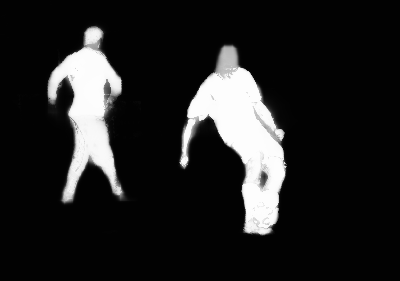} & \includegraphics[width=0.08\textwidth]{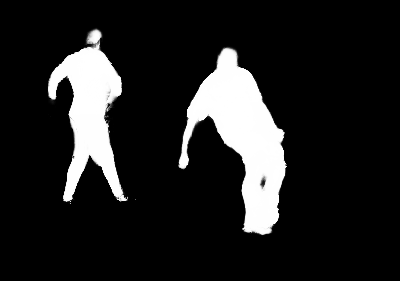} & \includegraphics [width=0.08\textwidth]{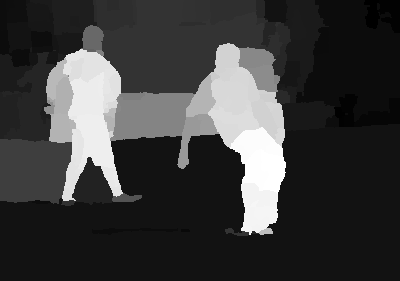} & \includegraphics [width=0.08\textwidth]{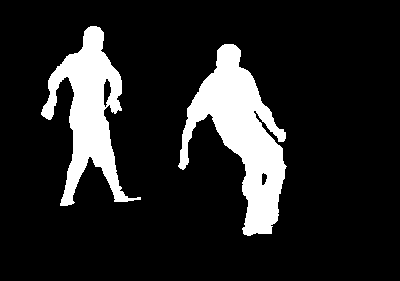} \\ 
\includegraphics[width=0.08\textwidth]{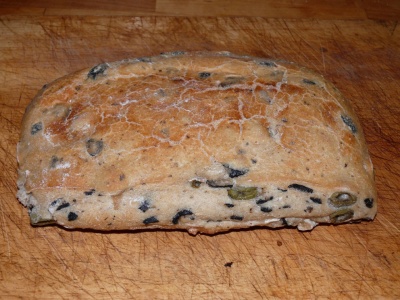} & \includegraphics [width=0.08\textwidth]{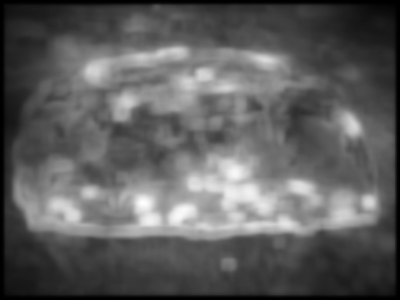} & \includegraphics [width=0.08\textwidth]{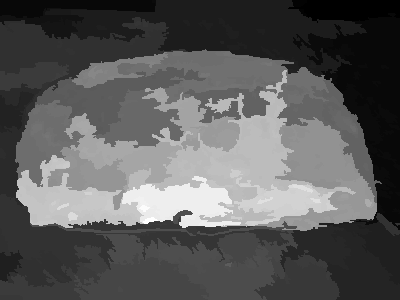} & \includegraphics [width=0.08\textwidth]{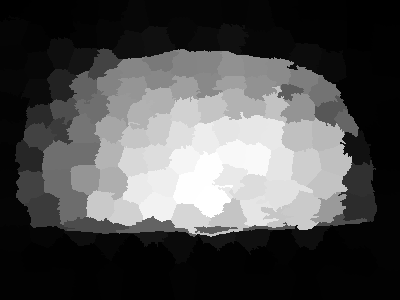}  &\includegraphics[width=0.08\textwidth]{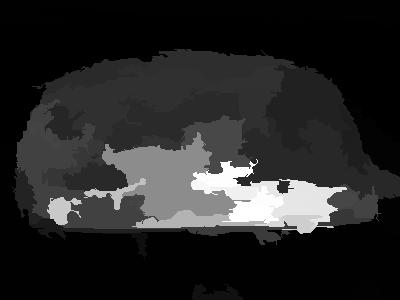} & \includegraphics [width=0.08\textwidth]{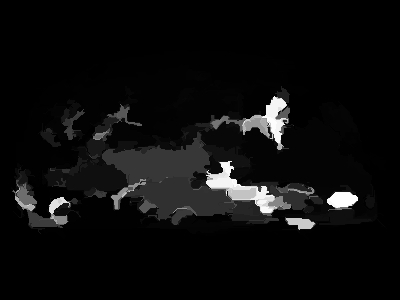} & \includegraphics [width=0.08\textwidth]{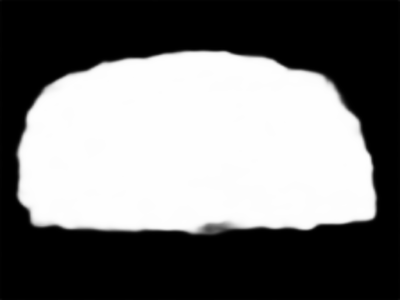} & \includegraphics [width=0.08\textwidth]{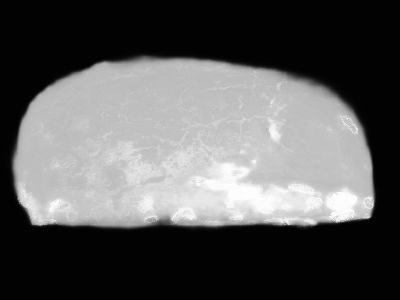} & \includegraphics[width=0.08\textwidth]{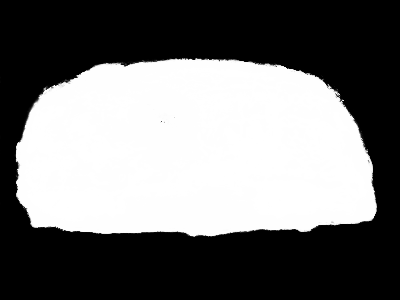} & \includegraphics [width=0.08\textwidth]{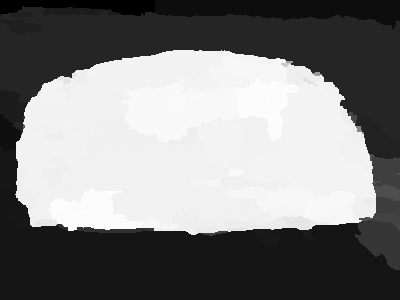} & \includegraphics [width=0.08\textwidth]{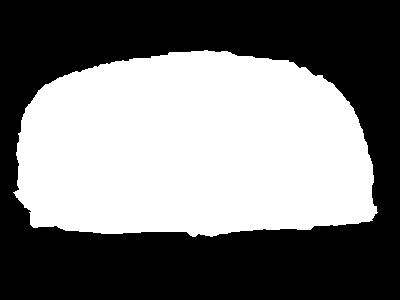} \\ 
\includegraphics[width=0.08\textwidth]{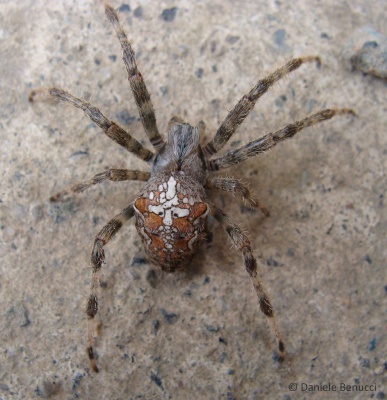} & \includegraphics [width=0.08\textwidth]{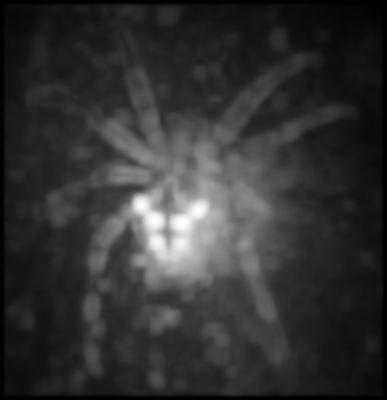} & \includegraphics [width=0.08\textwidth]{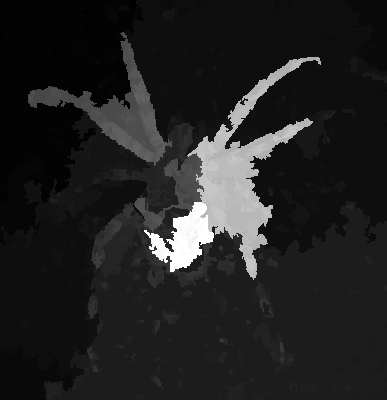} & \includegraphics [width=0.08\textwidth]{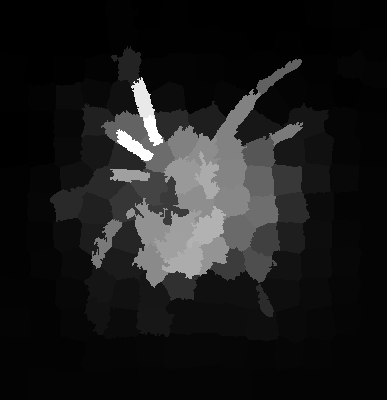}  &\includegraphics[width=0.08\textwidth]{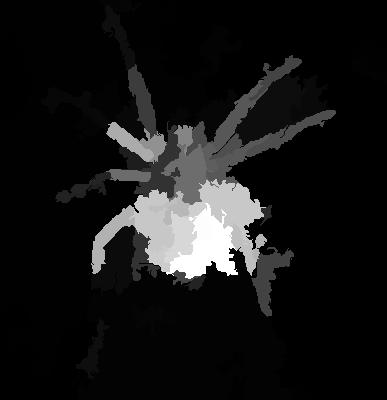} & \includegraphics [width=0.08\textwidth]{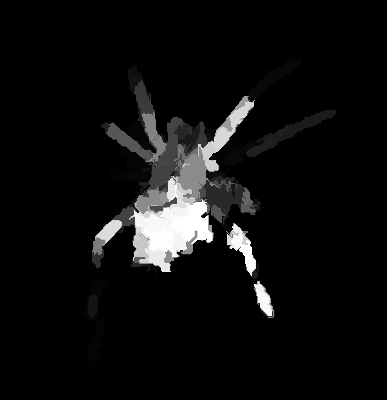} & \includegraphics [width=0.08\textwidth]{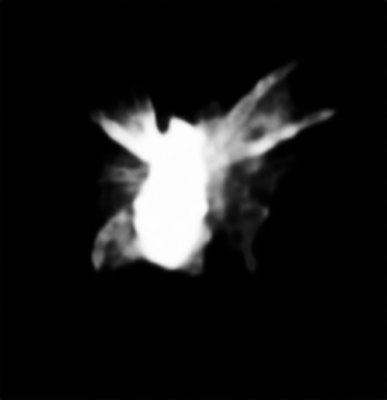} & \includegraphics [width=0.08\textwidth]{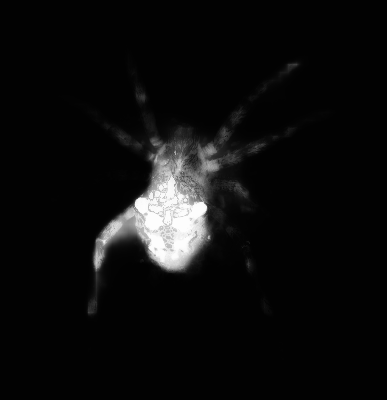} & \includegraphics[width=0.08\textwidth]{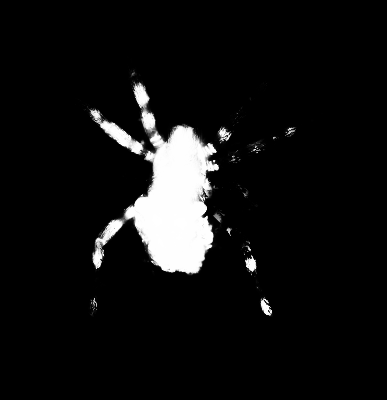} & \includegraphics [width=0.08\textwidth]{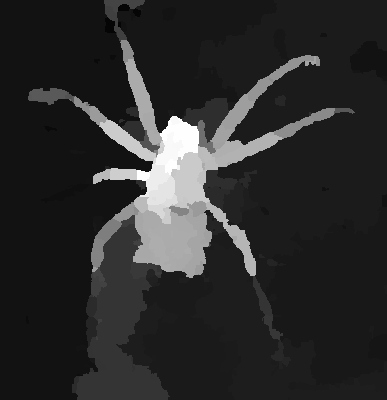} & \includegraphics [width=0.08\textwidth]{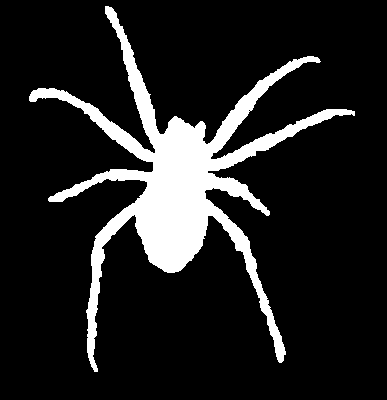} \\        
\includegraphics[width=0.08\textwidth]{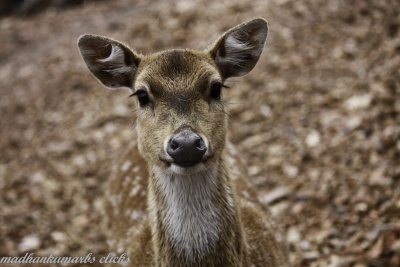} & \includegraphics [width=0.08\textwidth]{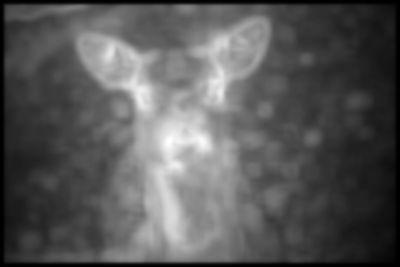} & \includegraphics [width=0.08\textwidth]{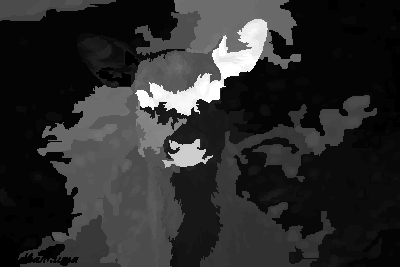} & \includegraphics [width=0.08\textwidth]{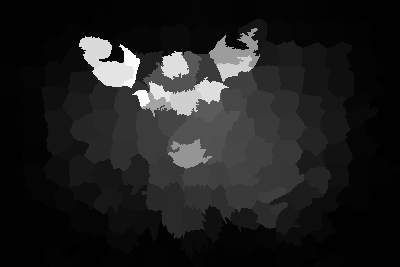}  &\includegraphics[width=0.08\textwidth]{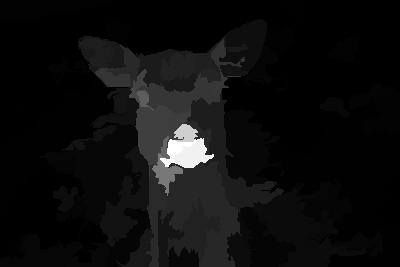} & \includegraphics [width=0.08\textwidth]{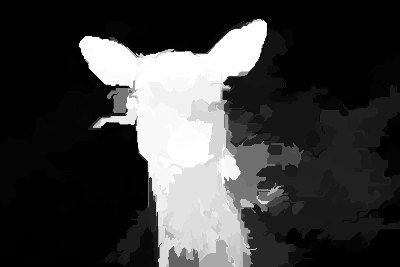} & \includegraphics [width=0.08\textwidth]{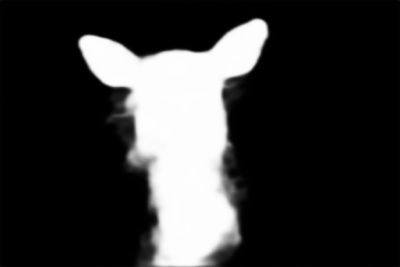} & \includegraphics [width=0.08\textwidth]{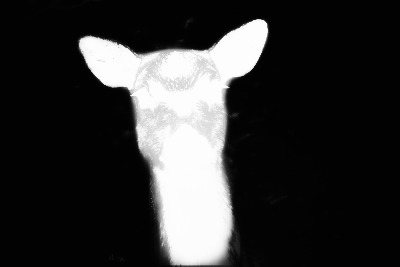} & \includegraphics[width=0.08\textwidth]{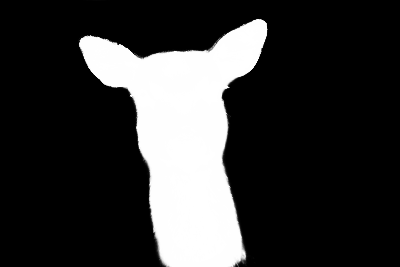} & \includegraphics [width=0.08\textwidth]{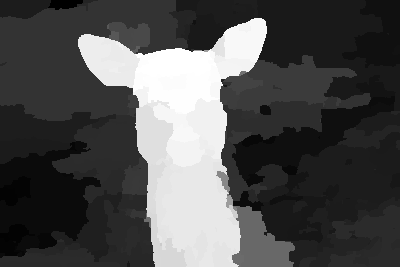} & \includegraphics [width=0.08\textwidth]{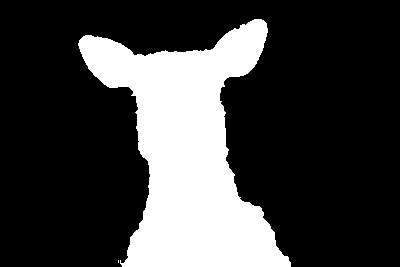} \\ 
\includegraphics[width=0.08\textwidth]{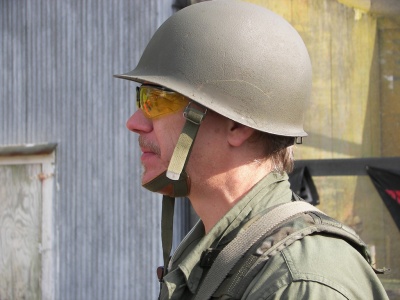} & \includegraphics [width=0.08\textwidth]{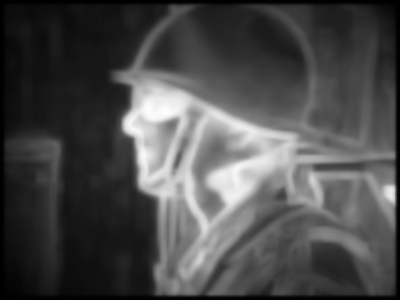} & \includegraphics [width=0.08\textwidth]{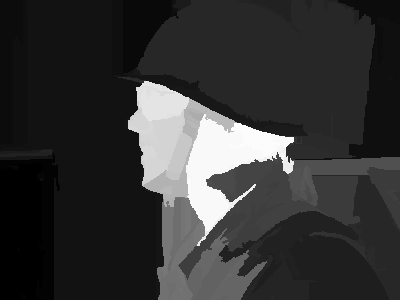} & \includegraphics [width=0.08\textwidth]{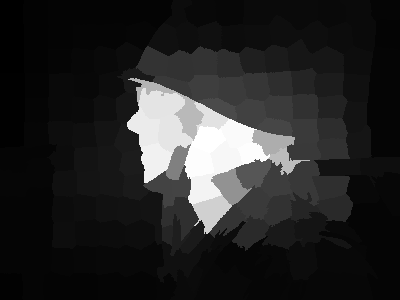}  &\includegraphics[width=0.08\textwidth]{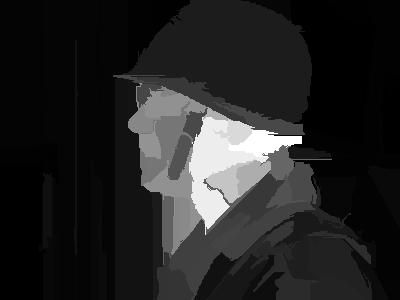} & \includegraphics [width=0.08\textwidth]{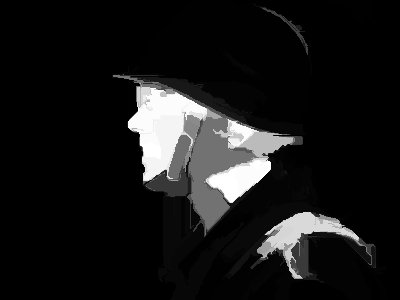} & \includegraphics [width=0.08\textwidth]{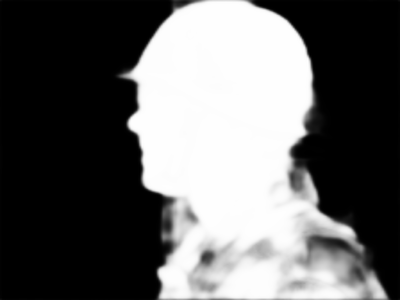} & \includegraphics [width=0.08\textwidth]{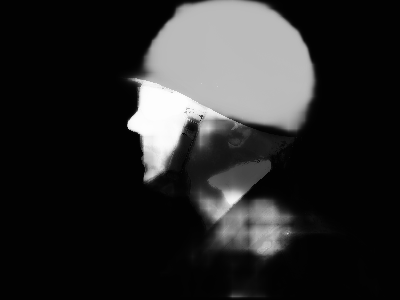} & \includegraphics[width=0.08\textwidth]{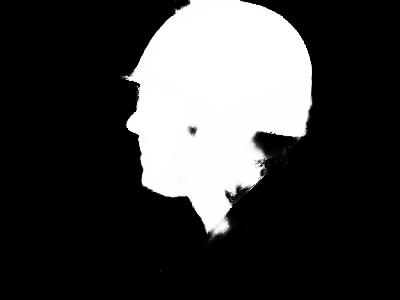} & \includegraphics [width=0.08\textwidth]{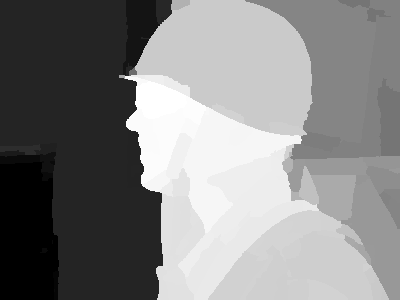} & \includegraphics [width=0.08\textwidth]{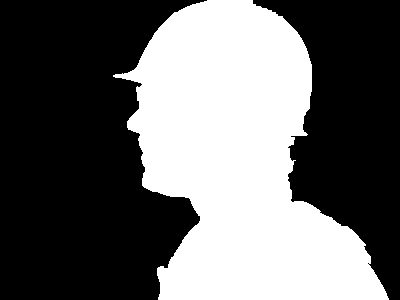} \\ 
\centering \tiny Image & \centering \tiny GOF & \centering \tiny HS & \centering \tiny GBMR & \tiny\centering DRFI & \tiny\centering MDF & \centering  \tiny DHS &  \centering \tiny DCL & \centering \tiny DSS & \centering \tiny Ours & \centering \tiny GT
\end{tabularx}
\caption{Visual comparison of saliency maps generated from 9 different methods, including our method. Methods for comparison includes DSS \cite{hou2017deeply}, DCL \cite{Li2016deep}, DHS \cite{liu2016dhsnet}, MDF\cite{li2015visual}, DRFI \cite{Jiang2013salient}, GOF \cite{goferman2012context}, HS \cite{yan2013hierarchical}, and  GBMR \cite{yang2013saliency}.}
\label{fig:examples}
\end{figure*}

We added a qualitative comparison in Fig.~\ref{fig:examples}. We tested our method in different challenging cases, multiple disconnected salient objects (first two rows), and low contrast between object and background (third and fourth row). Notice that our method correctly manages to assign saliency to most parts of the spider legs. Finally, results of objects touching the image boundary are shown where our method successfully includes the parts that touch the border (last two rows).

\begin{table}[!htbp]
\centering
\begin{tabular}{|c|c|c|c|c|} \hline
 & Pascal-S& MSRA-B& FT& ECSSD\\ \hline
 FT\citep{Achanta2009}&  $54.2$  &  $69.4$   & $77.7$    &  $60.7$ \\  \hline
 PRC\citep{lou2014regional} & $57.8$ & $ 69.2$ & $74.3 $ & $63.1 $ \\  \hline
 GOF\citep{goferman2012context} & $59.5$  & $69.7 $ & $71.1 $ & $63.7 $ \\ \hline
 PCAS \citep{margolin2013makes}& $59.6 $ & $71.6 $ & $83.5 $ & $65$ \\ \hline
 TD \citep{scharfenberger2013statistical} & $62.8$ & $75.4 $ & $83.3$ & $68.9$ \\ \hline
 HS\citep{yan2013hierarchical}  &$63.9$ & $ 81.1$  & $81.9$ & $72.8$ \\     \hline
 GBMR \citep{yang2013saliency}& $65.6$ & $82.5$ & 91.6 & $69.7$\\ \hline
 DRFI\citep{Jiang2013salient} & $69.3$ & $84.5$ & $ 83.3$ & $ 78$ \\ \hline
 LEGS\citep{wang2015deep} &   $75.2$ & $87$ & $-$ & $82.5$  \\ \hline
 MC\citep{zhao2015saliency} &   $79.3$ & $-$ & $-$ & $73.2$  \\ \hline
 MDF\citep{li2015visual}& $76.8^*$ & $88.5$ & $-$ & $83.2^*$  \\ \hline
 MTDS \citep{li2016deepsaliency}& $81.8^*$ & $-$ &  $-$ &  $80.9^*$ \\  \hline
 DHS\citep{liu2016dhsnet}& $82^*$ &  $-$  &  $-$  & $\textbf{90.5}^*$ \\  \hline
 DCL \citep{Li2016deep}&  $82.2^*$ &  $ 91.6$& $-$ & $89.8^*$ \\ \hline
 RFCN\citep{wang2016saliency}&   $\textbf{82.7}^*$ & $\textbf{92.6}$  & $-$ & $89.8 ^*$  \\  \hline
 DSS\citep{hou2017deeply} &  $\textbf{83}^*$ &  $\textbf{92.7}$  & $-$ & $\textbf{91.5}^*$ \\  \hline \hline
 Ours (trained on original trainset) & 82.3  & 90.9 & 91.5 & 86.9 \\ \hline
 Ours (trained on MSRA-B)& $78.1^*$ & 90.9 &$\textbf{91.8}^*$ &  $85.4^*$\\
 \hline
\end{tabular} 
\caption{Comparison of our method and context features against state-of-the-art methods. The results are based on training on the original trainset of each datasets. The methods which use the MSRA-B dataset to train are indicated with a $*$.}
\label{tab:comp}
\end{table}

\section{Conclusions}\label{sec:concl}
The direct context of an object is believed to be important for the saliency humans attribute to it. To model this directly in object proposal based saliency detection, we pair each object proposal with a context proposal. We propose several features to compute the saliency of the object based on its context; including features based on omni-directional and horizontal context continuity. 

We evaluate several object proposal methods for the task of saliency segmentation and find that multiscale combinatorial grouping outperforms selective search, geodesic object, SharpMask and Fastmask. We evaluate three off-the-shelf deep features networks and found that VGG-19 obtained the best results for saliency estimation. In the evaluation on four benchmark datasets we match results on the FT datasets and obtain competitive results on three datasets (PASCAL-S, MSRA-B and ECSSD). When only considering methods which are trained on the training set provided with the dataset, we obtain state-of-the-art on PASCAL-S and ECSSD.

For future research, we are interested in designing an end-to-end network which can predict both object and context proposals and extract their features. We are also interested in evaluating the usage of context proposals for other fields where object proposals are used, notably in semantic image segmentation. Finally, extending the theory to object proposals and saliency detection in video would be interesting \citep{rudoy2013learning}.

\section*{Acknowledgements}
This work has been supported by the project TIN2016-79717-R of the Spanish Ministry of Science, CHIST-ERA project PCIN-2015-226 and Masana acknowledges 2017FI-B-00218 grant of Generalitat de Catalunya. We also acknowledge the generous GPU support from NVIDIA corporation. 






\bibliographystyle{elsarticle-harv} 
\bibliography{mybibfile}

\end{document}